\pdfoutput=1

\documentclass[11pt,dvipsnames]{article}

\usepackage[]{acl}

\usepackage{times}
\usepackage{latexsym}
\usepackage{tabularx}
\usepackage{amsmath}

\usepackage[T1]{fontenc}

\usepackage[utf8]{inputenc}

\usepackage{microtype}

\usepackage{inconsolata}

\usepackage{graphicx}

\usepackage{booktabs}
\usepackage{multirow}
\usepackage{array}
\usepackage{caption}
\usepackage{siunitx}
\usepackage{enumitem}

\NewDocumentCommand{\heng}
{ mO{} }{\textcolor{red}{\textsuperscript{\textit{Heng}}\textsf{\textbf{\small[#1]}}}}

\NewDocumentCommand{\yi}
{ mO{} }{\textcolor{blue}{\textsuperscript{\textit{Yi}}\textsf{\textbf{\small[#1]}}}}

\newcommand*\inlinelargeimage[1]{\raisebox{-0.15\baselineskip}{$\,$\includegraphics[height=0.9\baselineskip]{#1}$\,\,$}}
\newcommand{\emojic}{\inlinelargeimage{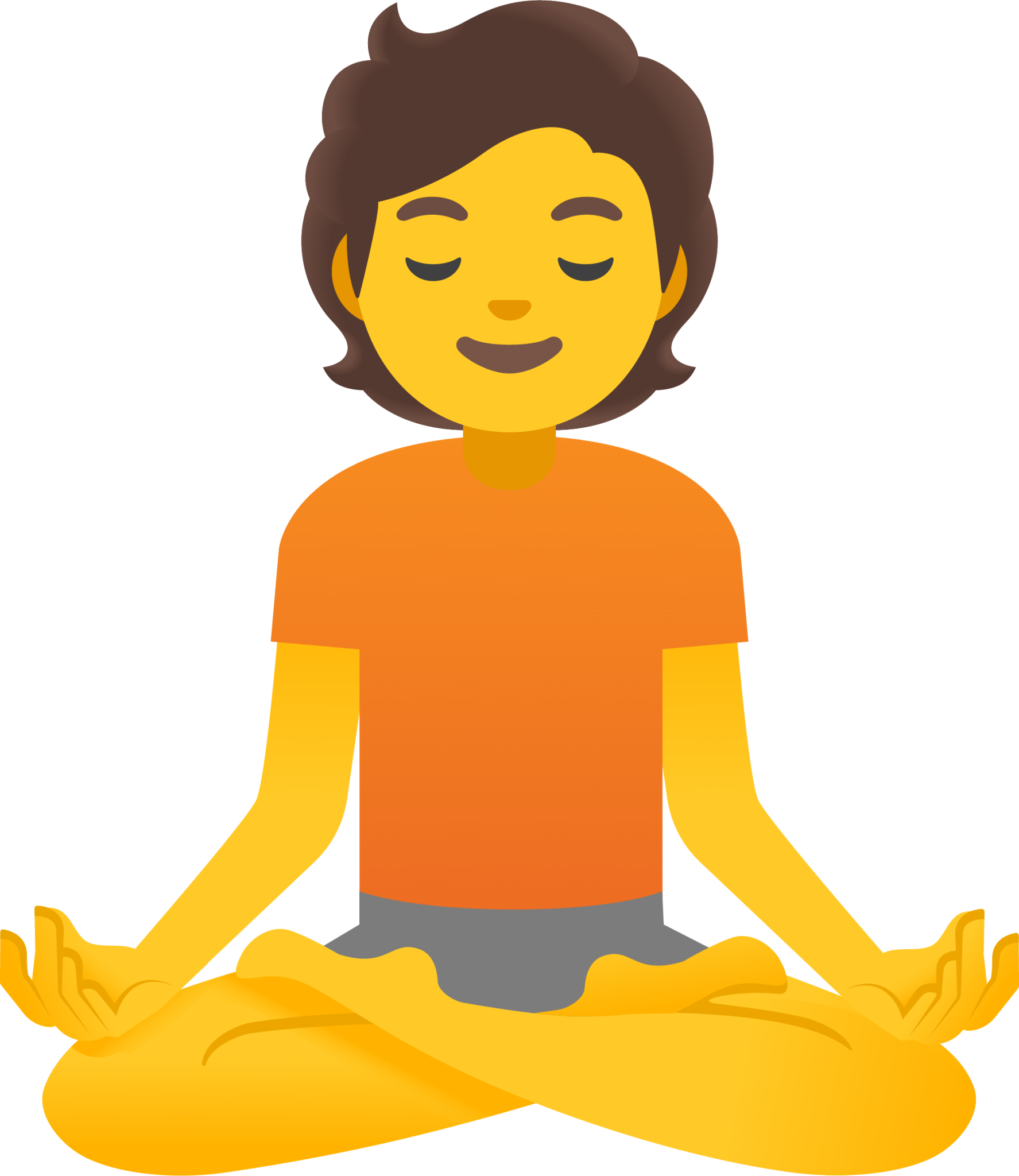}}

\title{\emojic CALM: Unleashing the Cross-Lingual Self-Aligning Ability of Language Model Question Answering
}

\author{~~Yumeng Wang$^{2}$ ~~~Zhiyuan Fan$^{2}$ ~~~Qingyun Wang$^{1}$ ~~\textbf{Yi R. (May) Fung}$^{2}\thanks{\ \  Corresponding author.}$\hspace{0.4em} \textbf{Heng Ji}$^{1*}$\\
$^{1}$University of Illinois Urbana-Champaign, ~~~~$^{2}$HKUST \\
\texttt{ywanglu@connect.ust.hk}  ~~~~~\texttt{yrfung@ust.hk}  ~~~~~\texttt{hengji@illinois.edu}
}  

\begin{document}
\maketitle

\begin{abstract}
Large Language Models (LLMs) are pretrained on extensive multilingual corpora to acquire both language-specific cultural knowledge and general knowledge. Ideally, while LLMs should provide consistent responses to culture-independent questions across languages, we observe significant performance disparities. To address this, we explore the \textbf{C}ross-Lingual Self-\textbf{A}ligning ability of \textbf{L}anguage \textbf{M}odels (\textbf{CALM}) to align knowledge across languages.
Specifically, for a given question, we sample multiple responses across different languages, and select the most self-consistent response as the target, leaving the remaining responses as negative examples. We then employ direct preference optimization (DPO) to align the model's knowledge across different languages.
Evaluations on the M{\small ED}QA and X-CSQA datasets demonstrate CALM's effectiveness in enhancing cross-lingual knowledge question answering, both in zero-shot and retrieval-augmented settings. We also found that increasing the number of languages involved in CALM training leads to higher accuracy and consistency. We offer a qualitative analysis of how cross-lingual consistency can enhance knowledge alignment and explore the method's generalizability\footnote{The source code and data of this paper is available on \url{https://github.com/wangym2/CALM}.}.

\end{abstract}

\section{Introduction}

LLMs have been pre-trained on various knowledge domains in multiple languages, capturing extensive world knowledge \cite{yu2024kolacarefullybenchmarkingworld}. This knowledge can be either sociocultural-dependent \cite{sun-etal-2023-decoding,liu-etal-2025-propainsight} or sociocultural-independent \cite{tang-etal-2024-mimir,huang-etal-2024-lvlms}. Ideally, LLMs should deliver consistent responses to the sociocultural-independent questions. However, due to the imbalance of the pretraining data, such knowledge is not well-aligned \cite{qi2023crosslingualconsistencyfactualknowledge,xu2024surveymultilinguallargelanguage,wu-etal-2025-aligning}. Research indicates that LLMs exhibit varying proficiency when addressing the same task across different languages \cite{xu2024surveymultilinguallargelanguage,huang20241+}. This variability stems from the difficulty of accessing knowledge encoded in one language while using others.

\begin{figure}
    \centering
    \includegraphics[width=1\linewidth]{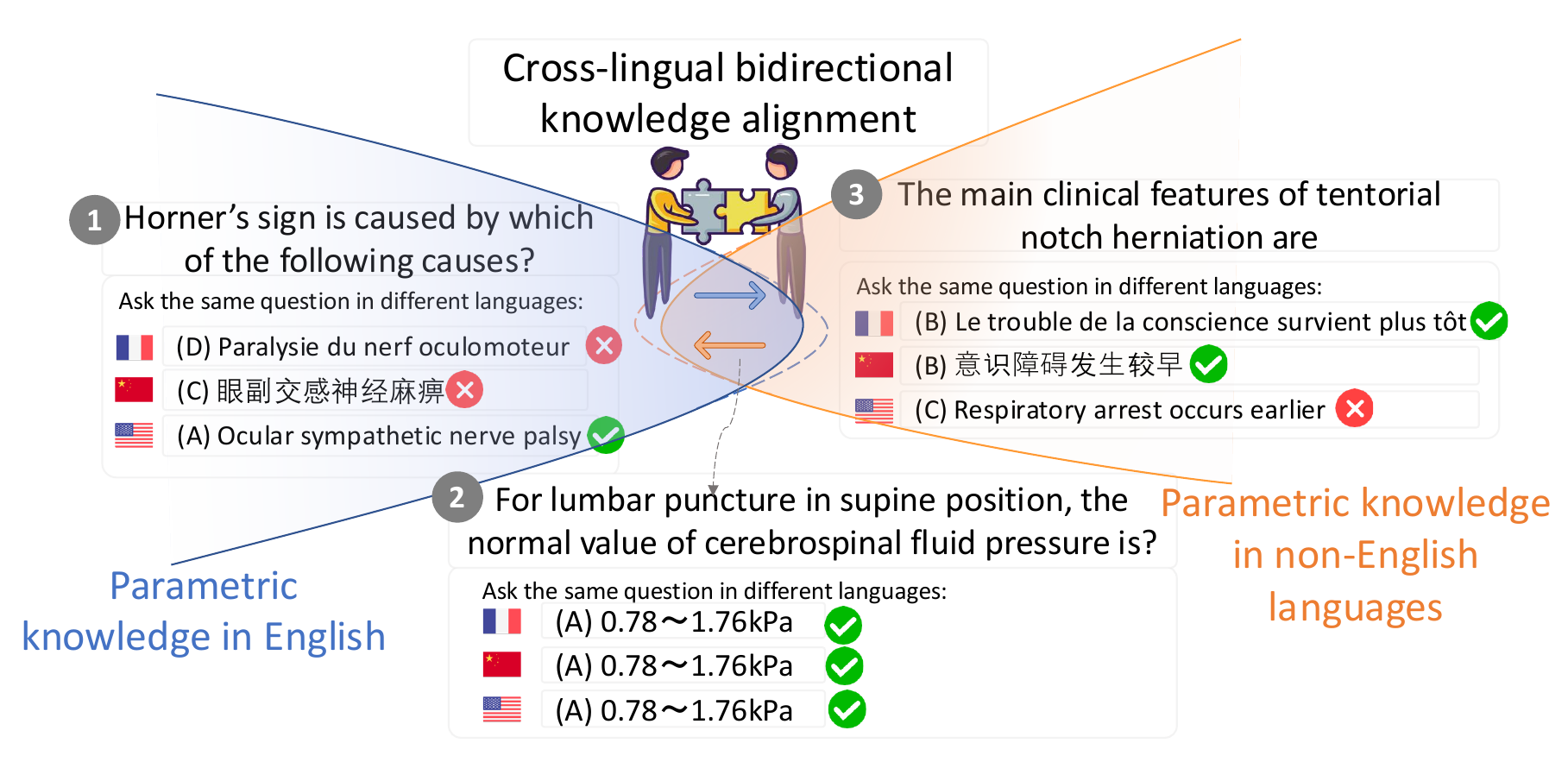}
    \caption{Knowledge is not well-aligned across languages. (1) represents knowledge encoded in English that is difficult to retrieve from other languages. (2) is the knowledge that is already well-aligned across languages. (3) is the knowledge encoded in other languages that is difficult to retrieve in English. Ideally, we want all the culture-independent knowledge to fall into (2).}
    \label{fig:fig1}
\end{figure}

To bridge the gap, recent papers introduced cross-lingual consistency \cite{qi2023crosslingualconsistencyfactualknowledge}, which pertains to the capacity to provide consistent responses across different languages when presented with the same query. The ultimate goal is to achieve language-agnostic question-answering proficiency in LLMs, enabling them to generalize effectively in multilingual environments. \citet{gao2024multilingualpretraininginstructiontuning} highlighted the positive impact of multilingual pretraining and instruction tuning on enhancing cross-lingual consistency. However, it also pointed out that current LLMs still face challenges in scaling up to improve cross-lingual knowledge retrieval capabilities.
\begin{figure*}
    \centering
    \includegraphics[width=1\linewidth]{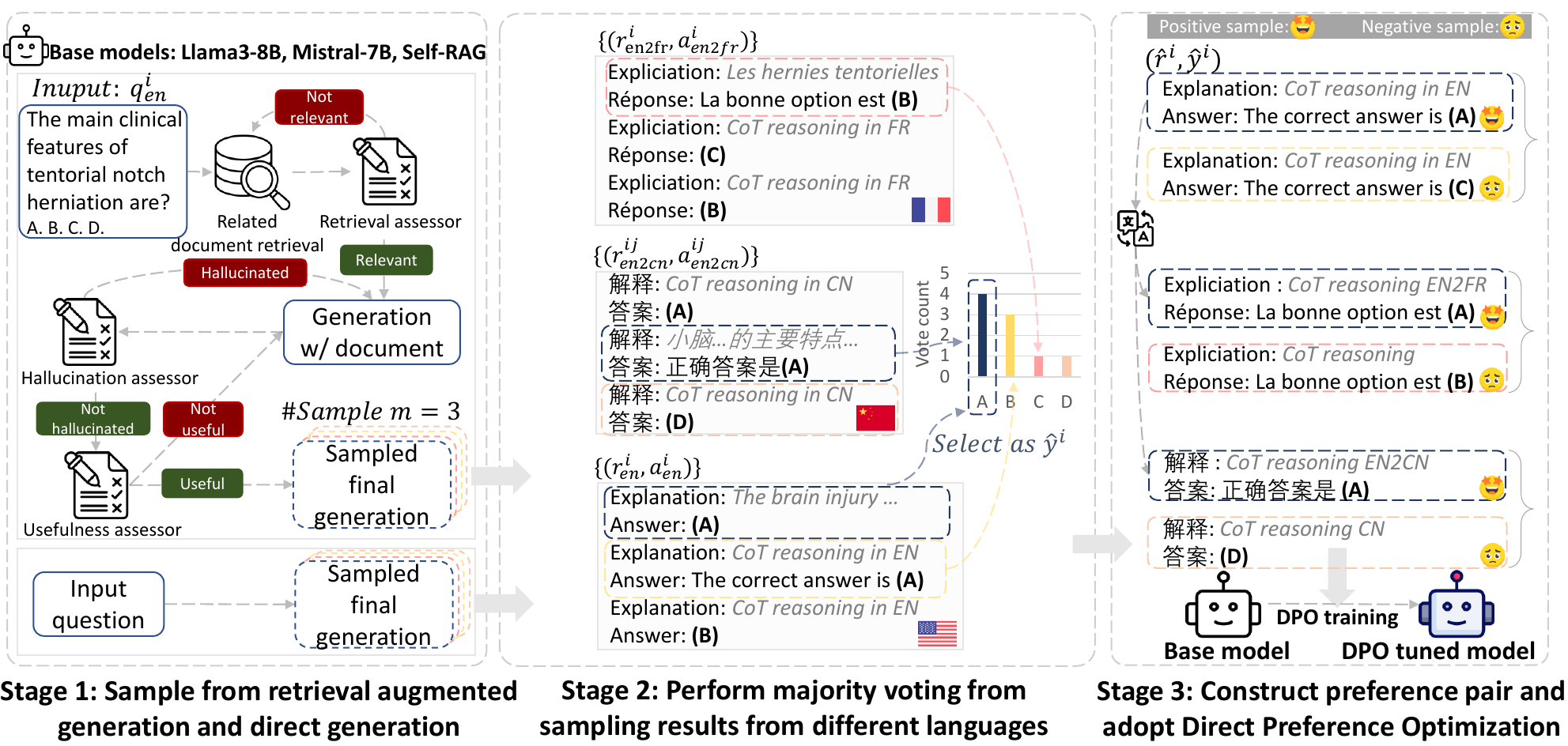}
    \caption{An example of the three stages in our proposed method assuming a question input originally in English.}
    \label{fig:fig2}
\end{figure*}
\citet{chen2023breakinglanguagebarriersmultilingual} utilized translation to develop a multilingual math reasoning instruction dataset. However, the challenge lies in the labor-intensive nature of obtaining high-quality translations and annotating data. \citet{she2024mapoadvancingmultilingualreasoning} leveraged translation consistency as a reward model to align the reasoning processes in other languages with the dominant language. Nevertheless, this approach may diminish the diversity of knowledge or reasoning introduced by different languages. \citet{huang20241+} enhanced the multilingual culture commonsense reasoning by implementing a multi-agent framework to aggregate the knowledge from diverse languages. In this work, we focus on leveraging multilingual knowledge aggregation by adopting preference optimization for model tuning.

To address the challenges of (1) establishing a scalable framework for aligning culture-independent knowledge across different languages and (2) lacking high-quality annotated data for training, we propose CALM, a method that encourages consistent answers to the same questions in different languages, motivated by the observation (Figure \ref{fig:fig1}) that non-English languages often contain complementary knowledge missing in English outputs. In Figure \ref{fig:fig3}, majority-voted answers consistently outperform English-only responses, making them viable alignment targets despite occasional factual inaccuracies. Exclusively aligning all other languages to English fails to leverage the LLM’s full multilingual knowledge potential, whereas CALM’s language-agnostic voting mechanism synthesizes cross-lingual insights.

Our approach leverages direct preference optimization (DPO) \cite{rafailov2024directpreferenceoptimizationlanguage} to facilitate cross-lingual alignment. The approach involves three steps. First, we sample a variety of multilingual Chain-of-Thought (CoT) outputs from the models. Next, we conduct majority voting on the sampled outputs in different languages, selecting the answer with the highest vote as positive. Finally, we pair the positive sample with all other answers that are inconsistent with it, utilizing these pairs for DPO training.
Moreover, we expand this framework to integrate external knowledge by combining Self-supervised Retrieval-Augmented Generation (Self-RAG) \cite{asai2023selfraglearningretrievegenerate} with DPO.

We conduct experiments on the challenging M{\small ED}QA \cite{jin2020diseasedoespatienthave} and the multilingual X-CSQA \cite{lin-etal-2021-common} datasets, each representing general knowledge and commonsense knowledge. On average, CALM boosts the accuracy on M{\small ED}QA and X-CSQA by +3.76\% and +5.55\% respectively.
Our key contributions are summarized:

\begin{table*}[t]
  \small
  \centering
  \begin{tabularx}{\linewidth}{>{\arraybackslash\hsize=1.6\hsize}X>{\centering\arraybackslash\hsize=0.9\hsize}X>{\centering\arraybackslash\hsize=0.9\hsize}X>{\centering\arraybackslash\hsize=0.9\hsize}X>{\centering\arraybackslash\hsize=0.9\hsize}X>{\centering\arraybackslash\hsize=0.9\hsize}X>{\centering\arraybackslash\hsize=0.9\hsize}X>{\centering\arraybackslash\hsize=0.9\hsize}X>
  {\centering\arraybackslash\hsize=0.9\hsize}X>{\centering\arraybackslash\hsize=0.9\hsize}X>
  {\centering\arraybackslash\hsize=0.9\hsize}X>
  {\centering\arraybackslash\hsize=0.9\hsize}X>
  {\centering\arraybackslash\hsize=0.9\hsize}X}
    \toprule
    \multirow{ 2}{*}{\textbf{Model}}& \multicolumn{3}{c}{\textbf{M{\small ED}QA (\%)}} & \multicolumn{9}{c}{\textbf{X-CSQA (\%)}} \\
    \cmidrule(lr){2-4} \cmidrule(lr){5-13}
      &  EN &  ZH & $ACC_{avg}$ & EN  & ZH  & FR & IT & DE & JA & $ACC_{avg}$ & Consis & AC3\\
    \midrule
    Llama & 60.1 & 56.2 & 58.2 & 73.1 & 52.1 & 60.8 & 59.8 & 57.5 & 49.2 & 62.0 & 57.73 & 58.24 \\
    + SFT              & 62.4 & 57.1 & 59.8 & 73.8 & 53.2 & 62.3 & 60.0 & 59.8 & 51.0 & 63.1 & 59.67 & 60.82 \\
    + \textbf{CALM} & \textbf{63.5} & \textbf{59.5} & \textbf{61.5} & \textbf{74.1} & \textbf{57.6} & \textbf{65.0} & \textbf{64.7} & \textbf{60.9} & \textbf{53.6} & \textbf{64.8} & \textbf{61.13} & \textbf{61.70}\\
    \midrule
    Self-RAG & 62.6 & 57.1 & 59.9 & - & - & - & - & - & - & - & - & - \\
    + SFT    & 63.8 & 60.3 & 62.1 & - & - & - & - & - & - & -  & - & - \\
    + \textbf{CALM} & \textbf{64.7}  &\textbf{63.7} &\textbf{64.2} & -& - & - & - & - & - & - & - & - \\
    \midrule
    Mistral & 49.8 & 36.4 & 43.1 & 60.1 & 48.3 & 51.6 & 50.7 & 49.4 & 43.0 & 53.3 & 50.51 & 50.51 \\
    + SFT   & 50.3 & 37.9 & 44.1 & 67.7 & 48.8 & 53.7 & 56.6 & 55.6 & 44.1 & 56.7 & 53.27 & 53.83 \\
    + \textbf{CALM}& \textbf{52.9} & \textbf{38.5} & \textbf{45.7} &\textbf{68.1} & \textbf{56.8} & \textbf{56.8} & \textbf{57.7} & \textbf{58.6} & \textbf{50.5} & \textbf{60.6} & \textbf{57.27} & \textbf{57.67}\\
    \bottomrule

  \end{tabularx}
  \caption{Model accuracy percentage score 
  on the test set of M{\small ED}QA and X-CSQA in different languages. ``$ACC_{avg}$'' denotes the average traditional accuracy of all languages, which represents the overall level of domain knowledge of the model. 
  The bold text represents the best result in the given model. Note that there are no X-CSQA results for Self-RAG because there are no documents available for retrieval. The full result of M{\small ED}QA can be found in Table \ref{medqa-full}.}
  \label{tab1}
\end{table*}

\begin{itemize}[leftmargin=*]
\itemsep0em 
    \item We propose CALM, a label-free approach to effectively align the culture-independent knowledge by encouraging cross-lingual consistency, enabling the model to enhance its knowledge accuracy and consistency \cite{huang-etal-2023-large}.
    \item We conduct experiments in both zero-shot Chain-of-Thought and retrieval augmented settings, utilizing Llama3-8B-Instruct \cite{dubey2024llama}, Self-RAG \cite{asai2023selfraglearningretrievegenerate}, and Mistral-7B-Instruct-v0.2 \cite{jiang2023mistral7b}. The outcomes highlight the efficacy of our approach in aligning internal and external knowledge.
    \item We further evaluate the cross-language and cross-dataset generalizability of CALM, showcasing its robustness and scalability.
\end{itemize}



\section{Method}
To encourage cross-lingual consistency, CALM samples a variety of Chain-of-Thought (CoT) \cite{10.5555/3600270.3602070,10.5555/3600270.3601883} responses from different languages, and leverages
response consistency \cite{wang2023selfconsistencyimproveschainthought,wu2025plata} as the learning signal. By selecting the most voted response as the positive sample, we construct the preference pairs and adopt DPO to optimize the preference. 
As the winning response may be any language, we preserve the diverse knowledge from languages other than English. We verified our approach in a retrieval-augmented setting, showing that our approach boosts the multilingual transferability of both internal and external knowledge. The proposed framework is shown in Figure \ref{fig:fig2}. Our method comprises multilingual response sampling, self-consistency-based preference pair construction, and multilingual knowledge alignment. 

\subsection{Multilingual response sampling}
\label{sec:3-1}
\textbf{Translation} For monolingual dataset, where a series of multiple choice questions are provided in its primary language (e.g., English), denoted as $Q_{en} = \{q_{en}^i\}_{i=1}^{N}$, we first translate them into two additional languages, say Chinese ($Q_{en2cn}$) and French ($Q_{en2fr}$). For multilingual datasets, this translation step is omitted, and the parallel questions in different languages are utilized directly.\\
\textbf{CoT answer generation} We apply multiple path decoding with temperature \textit{T = 1} on each variant of the question $q_{*}^i$ for all $i=1,..., N$ and * be any language in $\{en, en2fr, en2cn\}$ to generate \textit{m} pairs of CoT explanations and answers $\{(r_{\text{*}}^{ij}, y_{\text{*}}^{ij})\}_{j = 1}^m$, where $y$ denotes one of the predicted choice (A, B, C,...). The model is instructed to output an ``Explanation'' followed by an ``Answer'' to conform with the CoT format \cite{10.5555/3600270.3602070}.

\begin{figure*}[ht]
    \centering
    \includegraphics[width=1\linewidth]{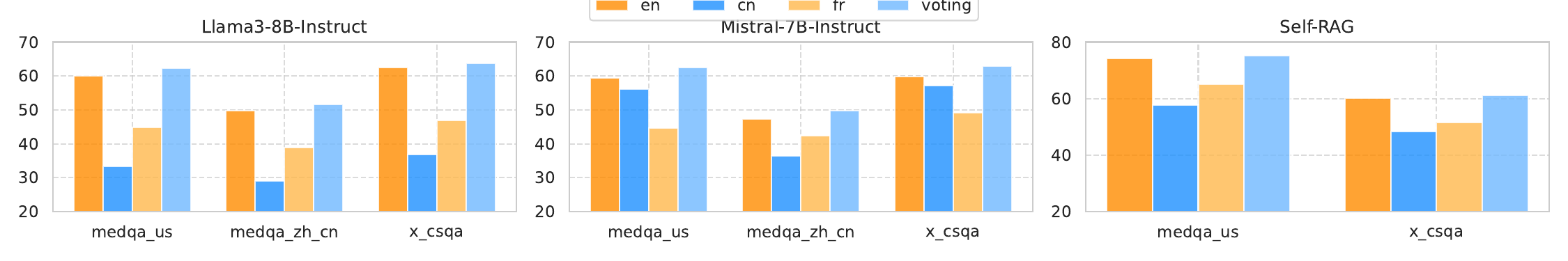}
    \caption{Visualization of mono-lingual (EN, ZH-CN, FR) percentage accuracy against the multilingual majority voting accuracy. The multilingual majority-voting result always has the highest accuracy. The proportion of each language in the CALM training data is in Table \ref{stats-train-portion}.}
    \label{fig:fig3}
\end{figure*}
\subsection{Self-consistency based preference pair construction}
\label{sec:3-2}
\textbf{Self-consistency} CALM assumes that the answer with the most votes reflects the highest model confidence \cite{xiong2024llmsexpressuncertaintyempirical,kabra2023programaidedreasonersbetterknow}, making it more likely to be correct \cite{wang2023selfconsistencyimproveschainthought}. use majority voting to identify the most popular option $\hat{y}_i$ from all multilingual answers, though $\hat{y}_i$ may not necessarily match the ground truth answer. We designate the most self-consistent answer as the positive sample.
\\
\textbf{Preference pair} After obtaining a set $S = \{(r^{ik}, y^{ik})\}_k$ of the most voted explanation-answer pair that satisfies $\forall y^{ik} \in \{(r^{ik}, y^{ik})\}_k, y^{ik} = \hat{y}^i$, we pair each of the positive samples with negative samples. Note that the positive samples are not necessarily in English. Hence, we aggregate the internal knowledge of both English and non-English languages. Negative samples are inconsistent with the positive ones, i.e., $y_{negative} \neq \hat{y}^i$. For each positive-negative sample pair, the positive sample is translated into the language of the negative sample. The final preference pairs of the i-th question are $p^i = \{{ p^i_w:(\hat{r}^i}_{trans}, \hat{y}^i), p^i_l:({r^i}, {y^i})_{neg}\}$.

\subsection{Multilingual knowledge alignment}
We adopt DPO as the alignment approach using the preference pairs $(p_w, p_l)$ obtained from \ref{sec:3-2}, where $p_w$ is preferred over $p_l$. Given an input question $q$, we optimize the following objective:

\begin{flalign*}
    & L_{\text{DPO}}(\pi_\theta;\pi_{ref}) = \mathbf{E}(q,p_w,p_l) \sim \nonumber \\
    & 
\mathcal{D}\left[\log\sigma\left(\beta\log\frac{\pi_\theta(p_w|q)}{\pi_{\text{ref}}(p_w|q)} -\beta\log\frac{\pi_\theta(p_l|q)}{\pi_{\text{ref}}(p_l|q)}\right)\right]
\end{flalign*}

\section{Experiment and Results}


\subsection{Datasets and Metrics}
\label{4-1}
We perform experiments on the following datasets:
\begin{itemize}[leftmargin=*]
    \itemsep0em 
    \item \textbf{M{\small ED}QA:} Zero-shot question answering, and Self-RAG's noisy evidence retrieval \cite{jin2020diseasedoespatienthave} over multiple evidence on medical multi-choice questions.
    \item \textbf{X-CSQA:} General multilingual commonsense question answering, including parallel questions from English, Chinese, French, Italian, German, and Japanese.
\end{itemize}
We adopt the multilingual consistency metrics introduced by \cite{wang2024seaevalmultilingualfoundationmodels,lin2024crossinefficientinstructiontuning}, which includes \textit{traditional accuracy}, \textit{consistency} and \textit{AC3}. Traditional accuracy refers to the accuracy of the multiple-choice questions. \textit{Consistency} is intended to measure if the model delivers consistent responses to the same question in different languages. A higher consistency score implies that multilingual LLMs can provide consistent responses across languages, which is irrelevant to the accuracy. For datasets like X-CSQA that contains a set of questions $Q = \{q^i\}_{i=1}^N$ across six languages, the consistency metric is defined as:
\begin{flalign}
    \notag M_{\{l_1,...,l_s\}} = \frac{\sum^{N}_{i=1} 1\{y^{l_1}_i=y^{l_2}_i=...=y^{l_s}_i\}}{N}
\end{flalign}
in which $y^{l_s}_i$ denotes the answer to the i-th multiple choice question given by language $l_s$. 
The final multilingual consistency is given by:
\begin{equation*}
\label{eq3}
   \notag Consistency_s = \frac{\sum_{\{l_1,l_2,...,l_s \in C(a,q_i)\}}M_{\{l_1,l_2,...,l_s\}}}{C_6^s}
\end{equation*}
\textit{AC3} is a metric combining accuracy and cross-lingual consistency, which is more robust for this multilingual task. The formulation is given by:
\begin{flalign*}
\label{eq4}
  AC3_s = 2 \times \frac{Accuracy \times Consistency_s}{Accuracy + Consistency_s}
\end{flalign*}
By considering both accuracy and multilingual consistency, we can measure the knowledge gain and the cross-lingual consistency.
\subsection{Baselines}
\paragraph{Base models} Our experiments utilize three base models, including Llama3-8B-Instruct \cite{dubey2024llama}, Mistral-7B-Instruct-v0.2 \cite{jiang2023mistral7b}, and Self-RAG \cite{asai2023selfraglearningretrievegenerate}. The testing results from the first two models demonstrate the efficacy of our approach in aligning internal knowledge, while the result from the last model highlights its proficiency in aligning external knowledge. The primary baseline is the direct inference results from all base models.

\paragraph{Supervised finetuning on preferred samples} To prove the necessity of DPO in training, we adopt supervised fine-tuning (SFT) \cite{luong-manning-2015-stanford} on preferred samples, namely using the most voted answers as SFT labels.



\subsection{Results}
In Table \ref{tab1}, CALM has encouraged the model to produce more accurate and consistent answers in all settings, outperforming the base model and the supervised fine-tuned model under all settings. Notably, the performance gain in X-CSQA surpasses that of M{\small ED}QA, which is likely due to the involvement of more languages participating, thereby activating more internal knowledge.
Therefore, we can conclude that our approach has successfully facilitated the cross-lingual self-alignment.
\section{Discussion}
\subsection{Accuracy of the positive samples}
In Figure \ref{fig:fig3}, we observe that the most self-consistent answer does not always align with the factually correct answer. Although the self-consistent answer’s accuracy slightly surpasses monolingual accuracy, the improvement remains modest. This raises an important question regarding the effectiveness of noisy labels in CALM's training process. To better understand this phenomenon, we examine examples of the preference data generated by CALM in Table \ref{table:qualitative} in the Appendix. The example shows that, although the preferred data may be factually incorrect, it often demonstrates better context awareness, which can lead the model to generate more accurate answers.


\begin{table}[h]
  \centering
  \small
  \scalebox{0.92}{
  \begin{tabular}{lcccccccc}
    \toprule
    \multirow{2}{*}{Model} & \multicolumn{2}{c}{M{\small ED}QA} & \multicolumn{3}{c}{X-CSQA}\\
    \cmidrule(lr){2-3} \cmidrule(lr){4-6} 
      & EN &  ZH & EN & ZH  & FR  \\
    \midrule
    Llama3-SFT w/ GT & 62.5 & 58.8	& 73.5 & 53.8 & 63.8  \\
    Llama3-DPO w/ GT&62.5	&59.3 &74.0& 54.3&64.1 \\
    Self-RAG-SFT w/ GT&63.6 &62.3 &- &- &- \\
    Self-RAG-DPO w/ GT&64.5 &63.8 &- &- &- \\
    Mistral-SFT w/ GT& 50.9 &36.9 &73.0 &51.6 &60.1 \\
    Mistral-DPO w/ GT& 52.4 &38.1	&73.2 &51.8 &55.0 \\
   
    \bottomrule
  \end{tabular}
    }
  \caption{Two additional baselines: DPO and SFT with ground truth. In this setting, we only keep the portion of DPO and SFT data that are factually correct.}
  \label{tab:gt-sft-dpo}
\end{table}
\subsection{SFT and DPO with ground truth}
Using ground truth from X-CSQA and M{\small ED}QA, we evaluate supervised SFT and DPO, retaining only preference pairs and SFT data where positive samples match ground truth. In Table \ref{tab:gt-sft-dpo}, supervised methods do not significantly outperform CALM, suggesting that guiding the model toward more confident and self-consistent answers can achieve comparable correctness even without ground truth.

 \begin{table}[h]
  \centering
  \scalebox{0.85}{
  \begin{tabular}{lcccccccc}
    \toprule
   \multirow{2}{*}{Model} & \multicolumn{3}{c}{M{\small ED}QA} & \multicolumn{2}{c}{X-CSQA}\\
    \cmidrule(lr){2-4} \cmidrule(lr){5-6} 
      & EN &  FR & ZH-CN & EN  & ZH-CN  \\
    \midrule
    Llama3-8B & 73.4 & 62.7 & 53.8 & 60.9 & 57.9  \\
    Mistral-7B & 70.8 &55.1 & 55.6 & 52.9 & 37.2  \\
   
    \bottomrule
  \end{tabular}
    }
  \caption{We investigate the cross-dataset generalizability. The table shows the result of training on M{\small ED}QA and testing on X-CSQA, or training on X-CSQA and testing on M{\small ED}QA. Both settings surpass the baseline.
}
  \label{cross-data-gen}
\end{table}
\subsection{Generalizability}
\paragraph{Cross-dataset generalizability}
To evaluate the generalizability, we conduct cross-dataset experiments by training models on X-CSQA and testing them on M{\small ED}QA, and vice versa. Table \ref{cross-data-gen} reveals that while the out-of-domain accuracy falls below the in-domain accuracy, it consistently exceeds the in-domain performance of the SFT baseline. This underscores the capability of CALM-trained models to provide multilingually consistent answers, even when faced with unseen tasks or domains. These findings suggest that CALM enhances in-domain performance and fosters robustness across different types of domains.

\paragraph{Cross-lingual generalizability} 
We implement CALM training sequentially, beginning with English and incrementally adding French and Chinese, progressing from high-resource to low-resource languages. At each step, we evaluate test accuracy across all languages. To assess CALM’s effectiveness in untrained languages, we include Japanese, Italian, and German in the test set, none of which were included during training. In Table \ref{cross-lingual-gen} in the Appendix, CALM demonstrates greater effectiveness as more languages participate in majority voting. Notably, even untrained languages exhibit accuracy improvements, suggesting that CALM’s alignment mechanism fosters a unified understanding of knowledge across languages, thereby enhancing overall comprehension. This aligns with \citet{she2024mapoadvancingmultilingualreasoning}, which similarly observe cross-lingual generalizability in multilingual reasoning tasks.

\section{Conclusion}
We introduce CALM, a novel framework to facilitate the alignment of LLM's knowledge across different languages. We observe that CALM is more effective when more languages are involved in the training, due to internal knowledge aggregation. Additionally, CALM outperforms ground truth DPO and SFT. It shows that although some of the positive samples are factually incorrect, they also contribute to the accuracy gain in CALM, possibly because more consistent answers often have better task understanding and can lead the model towards more correct answers. Through comprehensive experiments, we demonstrate the effectiveness of CALM in achieving robust cross-lingual knowledge alignment. 

\section*{Limitations}
One of the main limitations of our study is that due to the constraints of computational resources, we are unable to perform experiments on larger models. 
For the same reason, we are also not able to perform full-parameter fine-tuning and can only use LoRA DPO fine-tuning as an alternative. The translations in the experiment are done by Google Translate API, which may not be accurate sometimes because the dataset contains a many challenging medical terminology, hindering our final performance. For the DPO training data construction, since the accuracy after majority-voting is still low, the final alignment performance may be constrained by the noisy labels in the positive samples. Training one language after another can result in performance degradation in other languages. Future work can further investigate continual learning in multilingual knowledge alignment.
\section*{Ethics Statements}
In this paper, we present a method to align knowledge across multiple languages, ensuring equitable access to LLMs for users from diverse linguistic backgrounds. Our approach utilizes the model’s own outputs to perform cross-lingual alignment without the need for human annotations. By reducing dependence on manual labeling, this method enhances fairness, scalability, and inclusivity in multilingual AI, furthering the democratization of LLMs across global communities.
\bibliography{main}

\appendix

\begin{table*}[h]
\small
\begin{tabularx}{\textwidth}{lX}
\toprule
\small{Question}  & \small{The farmer could raise a good chicken, but he wasn't ready to \textbf{move up} to a what? A. rooster B. cow C. bought at store D. hen E. tuna} \textcolor{blue}{\textit{[Ground truth: B. cow]}}\\
\hline
        \small{Chosen}           &    \small{Explanation: The phrase ``move up" is an idiomatic expression that means to progress or advance to the next level or stage. In this context, it implies that the farmer has some experience with raising a chicken (a good one at that), but is \textbf{not yet ready to take on more complex or challenging tasks}. Answer: A. rooster} \textcolor{blue}{\textit{[Incorrect, but the reasoning is more plausible since it realized the answer should be a task more challenging than raising a chicken.]}}  \\ \hline
        \small{Rejected}        &   \small{Explanation: This question tests the development stage of a farmer. The farmer is now able to raise chickens, which shows that he has achieved certain experience and achievements. Well, if he's not ready to upgrade to something, only one of these options makes sense. Answer: D. Hen} \textcolor{blue}{\textit{[Incorrect, and it does not show the same level of task understanding as the Chosen one ]}} \\ 
\bottomrule

\end{tabularx}
    \caption{Qualitative example of CALM generated preference pair, where the chosen answer is not factually correct. The blue text shows the analysis. Although the chosen and rejected samples are both incorrect, the former pays better attention to the key part of the context ``\textbf{move up}" by mentioning that the farmer will be likely to face a more challenging task. This reasoning shows better context awareness and is more likely to lead to the correct answer.}
    \label{table:qualitative}
\end{table*}
\section{Training and inference configuration}
\label{sec:appendix}

We set $m = 3$ when sampling responses for each of the base models. We finally obtained 17244 and 2168 preference pairs from M{\small ED}QA and X-CSQA datasets, respectively. We used LoRA \cite{hu2021LoRAlowrankadaptationlarge} Fine-tuning method for DPO and SFT training. The training parameters are listed in Table \ref{table:train-dpo-sft}. The inference parameters are shown in Table \ref{table:inference}. All the experiments are performed on NVIDIA A100-SXM-80GB GPUs. We utilize the Llama3-8B-Instruct and Mistral-7B-Instruct model from LlamaFactory \cite{zheng2024llamafactory} framework for training and testing.

\begin{table}[h]
    \centering
    \scalebox{0.8}{
    \begin{tabular}{lll}
        \toprule
        \textbf{Parameter}                  &\textbf{DPO} &\textbf{SFT}\\ \hline
        Learning Rate                       &5e-6    &5e-5     \\ \hline
        num\_train\_epochs                  &3.0     &3.0\\ \hline
        lr\_scheduler\_type                 &cosine  &consine \\ \hline
        per\_device\_train\_batch\_size     &1       & 1\\ \hline
        warmup\_ratio                       &0.1     & 0 \\ \hline
        val\_size                           &0.06    & 0.06    \\ \hline
        pref\_beta                          &0.1     & -\\ \hline
        pref\_loss                          &sigmoid & -    \\ \hline
        per\_device\_eval\_size             &2       & 2 \\ \hline
        LoRA\_rank                          &8       & 8 \\ \hline
        LoRA\_alpha                         &16      & 16  \\ \hline
        LoRA\_trainable                     &q$_{proj}$,v$_{proj}$  &q$_{proj}$,v$_{proj}$ \\ \hline
        Optimizer                           &Adam    &Adam\\ \bottomrule
    \end{tabular}
    }
    \caption{DPO, SFT training parameter 
    }
     \label{table:train-dpo-sft}
\end{table}

\begin{table}[h]
    \centering
    \small
    \scalebox{1.0}{
    \begin{tabular}{ll}
        \toprule
        \textbf{Parameter}      &   \textbf{Value} \\ \hline
        Temperature           &   1     \\ \hline
        top\_p        &   0.9 \\ \hline
        max\_new\_tokens       &   512 \\ \hline
        per\_device\_eval\_batch\_size     &   4 \\ \bottomrule
    \end{tabular}
    }
    \caption{Model inference parameters}
    \label{table:inference}
\end{table}

\begin{table}[h]
  \centering
  \scalebox{0.85}{
  \begin{tabular}{lcccccccccc}
    \toprule
    \multirow{2}{*}{Model} & \multicolumn{3}{c}{M{\small ED}QA(\%)} & \multicolumn{3}{c}{X-CSQA(\%)}\\
    \cmidrule(lr){2-4} \cmidrule(lr){5-7} 
    & EN &  CN & FR & EN  & CN  & FR \\
    \midrule
    Llama3-8B & 58.2 & 17.1 & 24.8 & 52.9 &21.5 &25.6  \\
    Mistral-7B & 47.2 & 18.1 & 34.7 & 49.3 & 21.7 & 29.0  \\
   
    \bottomrule
  \end{tabular}
    }
  \caption{The percentages of positive samples for each language across task settings. English tasks up the largest portion of the positive samples, but there are also considerable amounts of Chinese and French samples. 
}
  \label{stats-train-portion}
\end{table}

\begin{table}[h]
\small
    \centering
    \scalebox{1.0}{
    \begin{tabular}{lccc}
        \toprule
                        &\textbf{EN}      &   \textbf{CN} &   \textbf{FR}\\ \hline
        M{\small ED}QA  &   21.4   &47.3  &31.3    \\ \hline
        Mistral        &20.5	&40.0	&39.5 \\ \bottomrule
    \end{tabular}
    }
    \caption{Percentage of Chinese, French and English language in final CALM training data.}
    \label{table:train-portion}
\end{table}

\begin{table*}[h]
  \centering
  \scalebox{0.8}{
  \begin{tabular}{lcccccccccccc}
    \toprule
    \multirow{2}{*}{Model} & \multicolumn{4}{c}{M{\small ED}QA US} & \multicolumn{4}{c}{M{\small ED}QA CN-ZH}\\
    \cmidrule(lr){2-5} \cmidrule(lr){6-9} 
      & Native EN &  EN2CN & EN2FR &AVG & Native CN  & CN2EN  & CN2FR  &AVG \\
    \midrule
    Llama3-8B-Instruct & 60.1 & 33.3 & 44.9 & 46.1 & 56.2 & 59.5 & 44.7 & 53.5 \\
    + SFT              & 62.4 \small{$\uparrow2.3$} & 36.1 \small{$\uparrow2.8$} & 45.8 \small{$\uparrow0.9$}& 47.8 \small{$\uparrow1.7$}& 57.1 \small{$\uparrow0.9$}& 59.9 \small{$\uparrow0.4$}& 46.4 \small{$\uparrow1.7$}& 54.5 \small{$\uparrow1.0$}\\
    + \textbf{CALM} & \textbf{63.5 \small{$\uparrow3.4$}} & \textbf{39.8 \small{$\uparrow6.5^*$}} & \textbf{46.3 \small{$\uparrow1.4$}}& \textbf{49.9 \small{$\uparrow3.8$}} & \textbf{59.5 \small{$\uparrow3.3$}} & \textbf{60.8 \small{$\uparrow1.3$}} & \textbf{47.4 \small{$\uparrow2.7$}} & \textbf{55.9 \small{$\uparrow2.4$}} \\
    \midrule
    Self-RAG & 62.6 & 36.8 & 46.9 & 48.8 & 57.1 & 59.8 & 49.1 & 55.3  \\
    + SFT    & 63.8 \small{$\uparrow1.2$} & 40.3 \small{$\uparrow3.5$}& 47.4 \small{$\uparrow0.5$}& 50.5 \small{$\uparrow0.7$}& 60.3 \small{$\uparrow3.2$}& 61.0 \small{$\uparrow1.2$}& 51.2 \small{$\uparrow3.7$}& 57.5 \small{$\uparrow2.2$}\\
    + \textbf{CALM} & \textbf{64.7 \small{$\uparrow2.1$}} & \textbf{42.6 \small{$\uparrow5.8$}} &\textbf{49.4 \small{$\uparrow2.5$}} & \textbf{52.3 \small{$\uparrow3.5$}} &\textbf{63.7 \small{$\uparrow6.6^*$}} & \textbf{64.3 \small{$\uparrow4.5$}} & \textbf{52.8 \small{$\uparrow3.7$}} &\textbf{60.3 \small{$\uparrow5.0$}}\\
    \midrule
    Mistral-7B-Instruct & 49.8 & 29.1 & 38.8 & 39.2 & 36.4& 47.3 & 42.4 & 42.0 \\
    + SFT        & 50.3 \small{$\uparrow0.5$} & 31.6 \small{$\uparrow2.5$}& 40.7 \small{$\uparrow1.9$}& 40.9 \small{$\uparrow1.7$}& 37.9 \small{$\uparrow1.5$}& 49.3 \small{$\uparrow2.0$}& 44.6 \small{$\uparrow2.2$}& 43.9 \small{$\uparrow1.9$}\\
    +\textbf{CALM}& \textbf{52.9 \small{$\uparrow3.1$}} & \textbf{32.7 \small{$\uparrow3.6$}} & \textbf{41.9 \small{$\uparrow3.1$}} & \textbf{42.5 \small{$\uparrow3.3$}} & \textbf{38.5 \small{$\uparrow2.1$}} & \textbf{51.8 \small{$\uparrow4.5$}} & \textbf{45.6 \small{$\uparrow3.2$}} & \textbf{45.3 \small{$\uparrow3.3$}} \\
    \bottomrule
  \end{tabular}
    }
  \caption{Full result on the translated M{\small ED}QA dataset.}
  \label{medqa-full}
\end{table*}

\begin{table*}[h]
  \centering
  \small
  \scalebox{1.0}{
  \begin{tabular}{lcccccc}
    \toprule
  
    Model  & EN &  FR & ZH-CN & IT  & DE & JA  \\
    \midrule
    Llama CALM w/ EN & \textbf{73.4}& 60.8&52.5&61.6&56.5&42.5 \\
    Llama CALM w/ EN+FR & \textbf{73.6}& \textbf{62.0}&52.4&62.0&56.2&43.6 \\
    Llama CALM w/ EN+FR+CN &\textbf{74.1} & \textbf{65.0} & \textbf{54.5} &62.3 &57.0 &44.0\\
    \bottomrule
  \end{tabular}
    }
  \caption{We investigate the cross-lingual generalizability by incrementally adding the training languages in CALM and observe the testing result on both trained and untrained languages. Here, in-domain languages (e.g. languages that appeared in the training data) are highlighted in bold font.
}
  \label{cross-lingual-gen}
\end{table*}

\section{Detailed use of the training dataset}
\subsection{Data source}
This section shows the details of the preliminary dataset selection in Section \ref{4-1}. 11.6k and 10k multiple choice questions were sampled from the M{\small ED}QA-ZH-CN and M{\small ED}QA-US question bank \cite{jin2020diseasedoespatienthave}. We also used all the Chinese and English textbooks provided by M{\small ED}QA to construct a vector database, which is necessary for the retrieval augmented generation. For X-CSQA \cite{lin-etal-2021-common}, we sampled 3k Chinese, English, and French questions.

\subsection{Statistics of the training datasets}
Table \ref{stats-train-portion} and Table \ref{table:train-portion} shows the percentages of positive samples for each language across task settings. English indeed tasks up the largest portion of the positive samples, but there are still considerable amounts of Chinese and French samples.
\subsection{Full result of M{\small ED}QA dataset}
For M{\small ED}QA, we first translate the native Chinese and English questions into other languages, forming a parallel training set in Chinese, English and French. The full testing result of the M{\small ED}QA is illustrated in Table \ref{medqa-full}. The accuracy is improved across all the languages after CALM tuning, and the native language has the largest performance gain. The performance of non-native languages is possibly constrained by the translation quality.  

\end{document}